\documentclass[letterpaper]{article}
\usepackage{iccc}
\usepackage{blindtext}

\usepackage{times}
\usepackage{helvet}
\usepackage{courier}
\usepackage{amsmath}
\usepackage{graphicx}
\usepackage{float}
\usepackage{booktabs} 

\title{Creative Transformation in Literary Texts: Modelling Change Across Representational Levels}
\author{Ioana-Roxana Boriceanu\textsuperscript{1}, Liviu P. Dinu\textsuperscript{1,2} \\
  \textsuperscript{1}Faculty of Mathematics and Computer Science \\
  \textsuperscript{2}Human Language Technologies Research Center \\
  University of Bucharest, Romania \\
  \texttt{ioana-roxana.boriceanu@s.unibuc.ro, ldinu@fmi.unibuc.ro} 
  \\}  %
\begin{document} 
\maketitle

%----------------------------------------------------------------------------------%
\begin{abstract}
Creativity is often framed as the production of novelty, yet many cultural works emerge through transformation of earlier artifacts and not through isolated invention. Drawing on theories of imitation by Gabriel Tarde and James Mark Baldwin, this paper models creativity as selective transformation across multiple levels of textual representation. We introduce a multi-level framework that compares literary texts across lexical, semantic, conceptual, structural, and narrative dimensions using directional alignment and control calibrated similarity measures. Applying the model to historically documented literary relationships, we show that different pairs preserve source structure at different representational levels while diverging in others. These transformation profiles provide a quantitative method for characterizing how imitation persists and where creative divergence occurs within literary works. 
\end{abstract}

%----------------------------------------------------------------------------------%
\section{Introduction}
Creativity has long been a central concern in psychology, philosophy, and the arts, and has gained renewed urgency in the context of artificial intelligence, where questions of human distinctiveness and creative capacity are increasingly foregrounded \cite{boden2016ai,marcus2019rebooting}. A widely accepted formulation is the standard definition proposed in \cite{runco2012standard}, which holds that creativity requires both originality and effectiveness. Novelty alone is insufficient, as arbitrary variation may be original without being meaningful, while usefulness without originality reduces to routine competence.

Margaret Boden further clarified the relation between novelty and value in \textit{The Creative Mind} \cite{boden2004creative}, distinguishing combinational, exploratory, and transformational creativity. In her account, creative production operates within structured spaces of possibility, and novelty emerges through transformations of those spaces instead of unconstrained deviation. Building on this perspective, computational creativity research has proposed formal frameworks for evaluating artificial artifacts \cite{maher2010evaluating,ritchie2007some,franceschelli2024creativity}. Many approaches measure novelty as distance from other artifacts within a class and value as performance relative to domain norms \cite{maher2010evaluating}. While these frameworks provide domain-independent evaluation criteria, they treat creativity as a property of an artifact relative to a distribution and not as a structured relation between specific works.

In many artistic domains, especially literature, new works do not simply occupy distant regions of a space of possibilities. They emerge through transformation of identifiable predecessors, retaining aspects of form, theme, or organization while reconfiguring others. Modelling creativity in such contexts therefore requires moving beyond class-level deviation toward an analysis of how representational structure is preserved and reshaped across works.

This relational view resonates with the social theory of Gabriel Tarde and the developmental psychology of James Mark Baldwin. In \textit{The Laws of Imitation} \cite{tarde1903laws}, Tarde argued that social phenomena propagate through imitation, with variations introduced in transmission becoming the basis of innovation. Baldwin similarly described imitation as a recursive process through which repetition consolidates habits while enabling modification \cite{baldwin1894imitation}. In both accounts, invention is not opposed to imitation but emerges from it. Novelty arises through selective transformation of inherited structure instead of originating from nothing.

Julia Kristeva, drawing on Bakhtinian dialogism, described every text as "a mosaic of quotations," framing intertextuality as the intersection of prior textual surfaces rather than the expression of a self-contained origin \cite{kristeva1980word}. Dialogism here refers to the idea that every utterance is shaped by and answers to prior and anticipated utterances \cite{bakhtin2010dialogic}. Roland Barthes likewise emphasized the plurality of cultural codes that converge within a work, shifting attention from authorial origin to the network of textual traces that constitute it \cite{barthes1977image}. Gérard Genette introduced analytic precision through his typology of transtextual relations, particularly his concept of hypertextuality, which foregrounds structured transformation between a later text and its precursor \cite{genette1997palimpsests}. Across these accounts, literary production appears not as isolated invention but as patterned reconfiguration of prior forms.

We operationalise this insight through a layered framework for textual comparison. Instead of collapsing two texts into a single similarity score, we decompose comparison across five representational strata: lexical surface patterns, semantic similarity in embedding space, conceptual category distributions, structural organization, and narrative dynamics across the text. This design draws on established traditions in stylistic analysis, stylometry, and influence detection \cite{burrows2002delta,reagan2016emotional,argamon2009automatically,burns2021profiling,xing2025modelling}, which have long examined these features as distinct dimensions of textual relation. This framework integrates these dimensions into a unified relational model oriented away from authorship attribution and class-level similarity and toward modelling selective transformation between specific works.

We further hypothesise that creative transformation is characterised by non-trivial retention combined with substantive divergence. If retention is negligible, the relation between works is indistinguishable from random variation; if retention is near total, the result approaches reproduction. Creativity therefore occupies a peaked region of the retention spectrum, contingent upon the presence of meaningful divergence in one or more other representational channels. It arises not at the extremes of imitation or rupture, but where structured inheritance coexists with structured transformation.

This paper makes three contributions:
\begin{enumerate}
    \item We introduce a multi-channel framework for modelling creative transformation between literary texts across lexical, semantic, conceptual, structural, and narrative levels.
    \item We propose a directional alignment and comparison pipeline that measures retention and divergence between texts while accounting for structural variation in narrative progression.
    \item We introduce a channelled aggregation mechanism that captures selective imitation by identifying the representational dimension in which structural inheritance is strongest while allowing other dimensions to diverge.
\end{enumerate}
The framework is intended as a diagnostic model of literary transformation, not as a definitive classifier of literary influence. It combines directional alignment, multi-channel representation, control-calibrated comparison, and channelled aggregation to produce interpretable transformation profiles for historically motivated reference pairs.

%----------------------------------------------------------------------------------%
\begin{table*}[t]
\centering
\small
\begin{tabular}{lllll}
\hline
Author & Work & Year & Literary Period & Tokens \\
\hline
Henry James & The Portrait of a Lady & 1881 & Late Victorian & 233k \\
Henry James & Daisy Miller & 1878 & Late Victorian & 22k \\
Edith Wharton & The House of Mirth & 1905 & American Realism & 134k \\
Edith Wharton & The Age of Innocence & 1920 & Modernism / Realism & 105k \\
George Eliot & Middlemarch & 1871 & Victorian Realism & 324k \\
George Eliot & The Mill on the Floss & 1860 & Victorian Realism & 214k \\
Jane Austen & Pride and Prejudice & 1813 & Regency & 123k \\
Jane Austen & Emma & 1815 & Regency & 165k \\
Thomas Hardy & Tess of the d'Urbervilles & 1891 & Late Victorian & 155k \\
Oscar Wilde & The Picture of Dorian Gray & 1890 & Aestheticism & 80k \\
Charles Dickens & David Copperfield & 1850 & Victorian & 364k \\
Charles Dickens & A Tale of Two Cities & 1859 & Victorian & 138k \\
Bram Stoker & Dracula & 1897 & Gothic / Late Victorian & 163k \\
Gustave Flaubert & Madame Bovary (Eleanor Marx trans.) & 1856 & French Realism & 118k \\
Gustave Flaubert & Salammbô (Eleanor Marx trans.) & 1862 & French Realism & 107k \\
James Joyce & Dubliners & 1914 & Modernism & 69k \\
James Joyce & Ulysses & 1920 & High Modernism & 270k \\
Homer & The Odyssey (Alexander Pope trans.) & 1725 & Neoclassical Translation & 115k \\
Homer & The Odyssey (Samuel Butler trans.) & 1900 & Victorian Translation & 119k \\
Homer & The Iliad (Pope translation) &  1715 & Classical Epic & 199k \\
Virginia Woolf & The Voyage Out & 1915 & Modernism & 140k \\
Virginia Woolf & Mrs Dalloway & 1925 & High Modernism & 65k \\
Mary Shelley & Frankenstein & 1818 & Romantic / Gothic & 75k \\
\hline
\end{tabular}
\caption{Corpus used in the experiments}
\label{tab:corpus}
\end{table*}

%----------------------------------------------------------------------------------%
\begin{table}[h]
\centering
\small
\begin{tabular}{ll}
\hline
Source Work & Target Work \\
\hline
The Portrait of a Lady & The House of Mirth \\
Pride and Prejudice & Middlemarch \\
Middlemarch & Tess of the d'Urbervilles \\
Odyssey (Pope) & Ulysses \\
Odyssey (Pope) & Mrs Dalloway \\
Odyssey (Butler) & Ulysses \\
Odyssey (Butler) & Mrs Dalloway \\
Odyssey (Pope) & Odyssey (Butler) \\
Madame Bovary & Dubliners \\
Madame Bovary & The Age of Innocence \\
\hline
\end{tabular}
\caption{Reference text pairs based on historically documented literary relationships}
\label{tab:influence}
\end{table}

%----------------------------------------------------------------------------------%
\section{Related Work}
This work draws on two research areas: computational models of creativity, and computational approaches to stylometry and literary influence.

\subsection{Computational Models of Creativity}
Computational creativity research has undergone several conceptual shifts. Early work formalized creativity as structured search within constrained conceptual spaces, modelling combinational, exploratory, and transformational movement over rule-governed domains \cite{boden2004creative,wiggins2006preliminary}. Creativity was therefore treated primarily as a generative process operating within a space of possibilities. Later work shifted attention toward evaluation, proposing empirical criteria and standardized procedures for assessing novelty and value in creative artifacts \cite{ritchie2007some,maher2010evaluating,jordanous2012standardised}. Creativity increasingly became a property to be measured as well as a process to be described.

With the rise of large-scale machine learning, a distributional perspective emerged. Generative systems learn statistical regularities and produce outputs that deviate from learned patterns \cite{franceschelli2022deepcreativity}. Novelty has been framed as prediction violation \cite{grace2019expectation}, as a balance between novelty and usefulness \cite{mukherjee2023creative}, or as a quantity difficult to stabilize across automatic metrics in large language models \cite{lu2025rethinking}. In parallel, feature-based approaches operationalize creativity as deviation within a corpus, computing scalar scores from similarity networks, semantic distance, or temporal drift in feature space \cite{shrivastava2017machine,orwig2025sweet,jimenez2022computational}.

\subsection{Stylometry, Authorship and Influence}
Computational stylometry is grounded in the premise that authors exhibit statistically stable stylistic patterns. Early quantitative studies showed that high frequency function words can reliably distinguish authors \cite{mosteller1963inference}. Later work expanded stylistic modelling to lexical, syntactic, and structural features such as n-grams, part-of-speech distributions, and vocabulary richness \cite{argamon2009automatically}. Distance-based measures, especially Burrows' Delta, further formalized stylistic similarity as deviation in frequency space \cite{burrows2002delta}.

With the adoption of machine learning, authorship attribution became increasingly framed as supervised classification. Stylometric features were combined with statistical and machine learning classifiers to scale across larger author sets and domains \cite{stamatatos2009survey}. Studies also examined the relative role of stylistic and topical signals \cite{sari2018topic}, while later approaches combined stylometric features with classical models \cite{alshamasi2022ensemble,boriceanu2025style} or used pretrained language models to detect stylometric differences \cite{barlas2020cross,fabien2020bertaa}.

Literary influence poses a more complex problem than authorship attribution. Instead of identifying the stylistic fingerprint of a single author, influence concerns relations between works produced by different authors and often separated by time, genre, or language. Early computational approaches focused on lexical overlap and text reuse, detecting quotation or phrase-level borrowing through n-gram similarity and related methods \cite{allen2011intertextuality,forstall2015modeling}. Later work moved beyond surface overlap through semantic representations, embeddings, topic models, and latent semantic analysis, allowing texts to be compared through thematic or latent similarity even when wording differs \cite{burns2021profiling,manjavacasa2020statistical,xing2025modelling,blei2003latent,jockers2013significant}. These approaches motivate the semantic channel of our framework, but our goal is not passage retrieval alone. We use semantic similarity as one component of a broader transformation profile.

Other work compares texts through stylistic, conceptual, and narrative structure. Function-word transition networks model style as relations between frequent grammatical markers \cite{segarra2015authorship}. WordNet supersenses and concreteness norms provide interpretable conceptual categories \cite{ciaramita2006broad,brysbaert2014concreteness}, while digital humanities work has compared narrative dynamics through sentiment trajectories, pacing, and plot progression \cite{reagan2016emotional}. These strands motivate the structural, conceptual, and narrative channels of the model.

Finally, influence may involve compression, expansion, omission, or rearrangement, making direct passage-to-passage comparison insufficient. Dynamic time warping compares ordered sequences that differ in length or pacing while preserving trajectory structure \cite{berndt1994using}. Prior studies have usually examined lexical reuse, semantic similarity, stylistic structure, conceptual categories, or narrative trajectories separately. Our framework combines these signals in a single directional, control-calibrated comparison of source and target texts, using the resulting five-channel profile to characterize creative transformation as selective retention and divergence.

%----------------------------------------------------------------------------------%
\section{Dataset}
The corpus consists of 23 literary works spanning the eighteenth to early twentieth centuries, including major texts from the English, American, and European literary canon. The dataset covers multiple literary periods, including Romanticism, Victorian realism, and high modernism, and contains approximately 3.5 million tokens in total. Table \ref{tab:corpus} lists the works included in the dataset together with their authors, publication dates, translators where applicable, literary periods, and approximate token counts after preprocessing.

The corpus was assembled to include both historically documented literary relationships and stylistically comparable but unrelated works. Instead of treating influence as a binary classification problem, the corpus serves as a comparative benchmark for analysing patterns of transformation between texts. Pairs of historically related works provide reference cases in which transformation is expected to occur, while unrelated comparisons provide a background distribution of stylistic similarity across the corpus.

To construct these reference comparisons, we define a set of text pairs based on well-established relationships discussed in literary scholarship. These include connections such as the influence of Henry James on Edith Wharton, Austen's influence on Eliot, Eliot's influence on Hardy, the influence of Flaubert on later realist and modernist writers, and the well-known engagement of Joyce and Woolf with Homeric epic. The dataset also includes comparisons between different translations of the \textit{Odyssey}, allowing the framework to capture relationships that preserve narrative structure while substantially altering lexical expression. The selected reference pairs are shown in Table \ref{tab:influence}.

In addition to these reference comparisons, control pairs are constructed by pairing texts from the corpus that are not historically linked through documented influence. These comparisons include both cross-tradition pairs and pairs drawn from similar literary periods. Together, these controls provide a background distribution of stylistic and structural similarity against which the transformation profiles of the reference pairs can be compared.

The texts were obtained from the Project Gutenberg digital library. Prior to analysis, non-content material was manually removed, including metadata such as source information, author descriptions, and editorial notes, retaining only the main body of the literary work. All texts are analysed in English. Non-English works are included through established English translations. The text was normalized through lowercasing, and tokenization was performed using a rule-based tokenizer.

For downstream comparison, texts were segmented into sequential units using one of two strategies: chapter boundaries or overlapping sliding windows. Overlapping sliding windows were constructed to reach a target size of approximately $2000$ tokens, with successive windows advanced by a fixed step of $500$ tokens. This segmentation allows structural patterns to be compared across works of different lengths while preserving local narrative organization.

%----------------------------------------------------------------------------------%
\section{Method: The Channelled Imitation Model}
To operationalise the theoretical framework proposed by Gabriel Tarde and James Mark Baldwin, our pipeline models literary creativity as structured transformation of prior texts. In this view, new works emerge through selective preservation and modification of earlier textual structures. Instead of treating creativity as deviation from a statistical distribution of artifacts, the model measures how representational structures are retained or transformed between specific works.

The proposed framework therefore compares a source text and a derived text across multiple representational layers. Each channel captures a different level of textual organisation. Higher similarity within a channel indicates structural inheritance, while divergence reflects creative transformation. By examining patterns of retention and divergence across channels simultaneously, the model characterises how literary works inherit and reshape earlier forms.

\subsection{Preprocessing and Directional Alignment}
Before influence can be analysed, the texts must first be placed into a comparable structure. Because derived works frequently omit, reorder, or expand upon material from their sources, direct one-to-one alignment between passages is insufficient. A later work may compress an episode, elaborate on a minor event, or introduce entirely new sections while still maintaining a recognisable structural relationship to the original.

To account for these transformations, each text is segmented into sequential windows, as described in the previous section, and encoded using Sentence-BERT embeddings (specifically the \texttt{all-MiniLM-L6-v2} model) \cite{reimers2019sentence}. Alignment between the two texts is then computed using Dynamic Time Warping (DTW) \cite{berndt1994using}, which allows flexible many-to-one or one-to-many correspondences between windows while preserving overall narrative order.

The alignment procedure is directional. Instead of treating the texts symmetrically, the algorithm anchors the derived text ($B$) to the source text ($A$), reflecting the theoretical assumption that influence flows from the earlier work toward the later one. To reduce matches between passages that occur at very different relative positions in the two texts, the DTW cost function incorporates a positional bias:
\begin{equation}
D(i,j)=d_{cosine}(A_i,B_j)+\lambda\left|\frac{i}{N}-\frac{j}{M}\right|
\end{equation}
where $d_{cosine}$ is the cosine distance between embedding vectors for windows $A_i$ and $B_j$, $N$ and $M$ denote the number of windows in the two texts, and $\lambda$ controls the strength of the positional bias. The second term measures the difference between the normalized positions of the two windows within their respective texts. Alignments between windows that occur at very different relative narrative positions therefore incur a higher cost. This biases the alignment toward passages that occupy comparable positions in the two works while still allowing local expansions or contractions of narrative material. In all reported experiments we set $\lambda = 0.15$. Because both the cosine-distance and positional terms are bounded in $[0,1]$, this caps the positional penalty at roughly $15\%$ of the maximum semantic cost, ensuring it acts as a tie-breaker rather than overriding semantic correspondence. To keep alignments locally monotonic, the DTW search is restricted to a band proportional to the length of the longer sequence, and the traceback is biased toward diagonal steps so that each source window receives a corresponding match.

%----------------------------------------------------------------------------------%
\begin{table}[t]
\centering
\small
\renewcommand{\arraystretch}{1.35}
\begin{tabular}{@{}l p{5.7cm}@{}}
\toprule
\textbf{Channel} & \textbf{Representative features} \\
\midrule
Lexical    & Token-frequency and content-word distributions, punctuation statistics, type--token ratio, inter-window lexical overlap \\
Structural & Function-word adjacency networks (density, degree distributions), part-of-speech transition patterns \\
Semantic   & Sentence-BERT embeddings, LDA topic mixtures, LSA representations \\
Conceptual & Supersense categories, named-entity-type distributions, lexical concreteness \\
Narrative  & Sentence-length pacing, sentiment trajectories, syntactic-complexity variation \\
\bottomrule
\end{tabular}
\caption{Feature layers used to capture each channel of textual transformation.}
\label{tab:features}
\end{table}

%----------------------------------------------------------------------------------%
\subsection{Multi-Level Representational Channels}
Once the texts are aligned, their relationship is analysed across five representational channels corresponding to different levels of textual organisation. Table \ref{tab:features} presents the feature layers extracted for each channel.

\textbf{The lexical channel} captures surface-level stylistic similarity. Each window is represented as a normalized token frequency distribution. Divergence between the source distribution ($Q$) and derived distribution ($P$) is measured using a dampened Kullback--Leibler divergence:

\begin{equation}
D_{KL}^{\text{damp}}(P \parallel Q)
=
\ln\left(1+\max(0,S)\right)
\end{equation}

\begin{equation}
S=\sum_x P(x)\ln\frac{P(x)}{Q(x)}
\end{equation}

This formulation measures lexical novelty while preventing extreme divergence values when the derived text introduces entirely new vocabulary.

\textbf{The structural channel} models stylistic organisation independent of topical content. Function word adjacency networks \cite{segarra2015authorship} are constructed from sequences of stop words within each window. Nodes correspond to function words, while edges represent transitions between successive function words in the text.

From these networks we compute structural metrics including network density and node degree distributions, capturing the global topology of function word usage. Structural similarity between aligned windows is evaluated using cosine similarity and Jensen--Shannon divergence over these distributions. Because function words are less topic dependent than content vocabulary, these measures provide a useful proxy for stylistic and structural inheritance.

\textbf{The semantic channel} captures thematic similarity through dense latent representations of meaning. Each window is encoded using contextual embeddings derived from Sentence-BERT, producing dense vector representations optimised for semantic similarity. Pairwise similarity between aligned windows is measured using cosine similarity between embedding vectors.

To complement embedding similarity, we also compute thematic similarity using topic modelling and latent semantic analysis (LSA). Topic distributions are obtained with Latent Dirichlet Allocation (LDA) using 50 topics, while LSA representations are derived from a truncated singular value decomposition of the term--document matrix with 100 dimensions. We use LDA because the corpus is relatively small and because topic mixtures provide an interpretable way to compare themes across aligned windows. The number of topics was fixed before evaluating the reference pairs, and the semantic channel combines LDA with other signals, so the results are not based on a single topic-model configuration. Together, these representations capture thematic similarity even when surface wording differs substantially.

\textbf{The conceptual channel} captures persistence of semantic categories that may remain stable even when lexical expression and sentence structure change. While the semantic channel models similarity in dense latent representations of meaning, the conceptual channel compares distributions of interpretable semantic categories across the text.

Three types of signals are used: lexical supersense categories \cite{ciaramita2006broad}, named entity type distributions, and lexical concreteness scores \cite{brysbaert2014concreteness}. These distributions provide a general representation of the conceptual landscape of a text, enabling the model to detect continuity of semantic domains even when vocabulary diverges substantially.

\textbf{The narrative channel} models large-scale narrative organisation by analysing how stylistic and emotional signals evolve across the text. This layer captures patterns that unfold across sequences of windows, including sentence-length based pacing, sentiment variation, and changes in structural complexity, instead of focusing only on individual sentences.

For each text, these signals are represented as time series defined over the ordered sequence of windows. For example, sentence-length statistics provide a proxy for narrative pacing, while sentiment scores capture shifts in emotional tone \cite{reagan2016emotional}. Similarity between the resulting narrative trajectories is then computed using correlation and cosine similarity between the corresponding time series. This representation allows the model to measure whether two texts exhibit similar narrative dynamics, such as comparable pacing patterns or sentiment arcs, even when the specific events or wording differ.

\subsection{Discriminative Feature Weighting}
Literary works produced within the same historical tradition often share stylistic conventions that do not reflect direct influence. To distinguish genuine textual inheritance from background similarity, the model applies a discriminative feature weighting procedure. For each feature $f$, a weight is computed as

\begin{equation}
w_f =
\max\left(
0.01,
\frac{\sigma_{tradition}(f)}
{\sigma_{global}(f)+\epsilon}
\right)
\end{equation}
where $\sigma_{tradition}$ measures feature variance among texts belonging to the same literary tradition, assigned by literary and national grouping such as English Victorian, French realist, or classical epic, and $\sigma_{global}$ measures variance across the entire corpus. The constant $\epsilon$ is a small positive value added for numerical stability to prevent division by zero when global variance is extremely small. Features that exhibit higher variance within a tradition relative to their global variance receive higher weight.

These weights are used to compute standardized feature scores relative to tradition-matched control pairs. The resulting weighted score provides a calibrated estimate of transformation-based influence.

\subsection{Channelled Aggregation of Influence}
The per-channel scores $z_c$ used in this section are computed from the discriminatively weighted features described above, standardized against tradition-matched control pairs. Because we hypothesise that creative transformation occurs selectively across representational levels, influence is not expected to appear uniformly across all channels. A derived work may preserve narrative structure while transforming lexical expression, or maintain conceptual themes while radically altering stylistic form.

Each representational channel therefore produces an influence signal measuring the degree to which that level of structure is retained between the aligned texts. Let $z_c$ denote the standardized similarity score for channel $c$. The resulting five-dimensional vector
\begin{equation}
\mathbf{z} =
(z_{narr}, z_{struct}, z_{sem}, z_{concept}, z_{lex})
\end{equation}
represents the influence profile between the two works.

The model identifies the strongest retained structure as the primary locus of inheritance:
\begin{equation}
z_{peak} = \max_c(z_c)
\end{equation}

Creative transformation often involves strong retention in one channel combined with divergence in others. To capture this pattern, we introduce a diagnostic scoring function that highlights channelled inheritance across representational levels:
\begin{equation}
I = \max(0, z_{peak}) (1 + \alpha n_{neg})
\label{eq:score}
\end{equation}
where $n_{neg}$ is the number of channels whose standardized score falls below a divergence threshold of $z_c < -0.5$, corresponding to half a standard deviation below the control mean, and $\alpha$ is a scaling constant set to $0.20$. The score is primarily determined by the strength of the dominant channel, with the second factor providing a bounded adjustment for the sharpness of the channelling pattern. Because at most four of the five channels can diverge while one is retained, the maximum bonus is $1 + 0.20 \times 4 = 1.80$, ensuring that the divergence term modulates the score without dominating it. This design reflects the theoretical expectation that creative transformation is characterised by a distinctive configuration in which strong inheritance in a specific representational dimension coexists with departure from the source in others, instead of by uniform similarity or uniform distance. The scoring function should therefore be interpreted as a diagnostic of selective transformation rather than as a measure of overall similarity between texts.

%----------------------------------------------------------------------------------%
\begin{figure}[t]
    \centering
    \includegraphics[width=\linewidth]{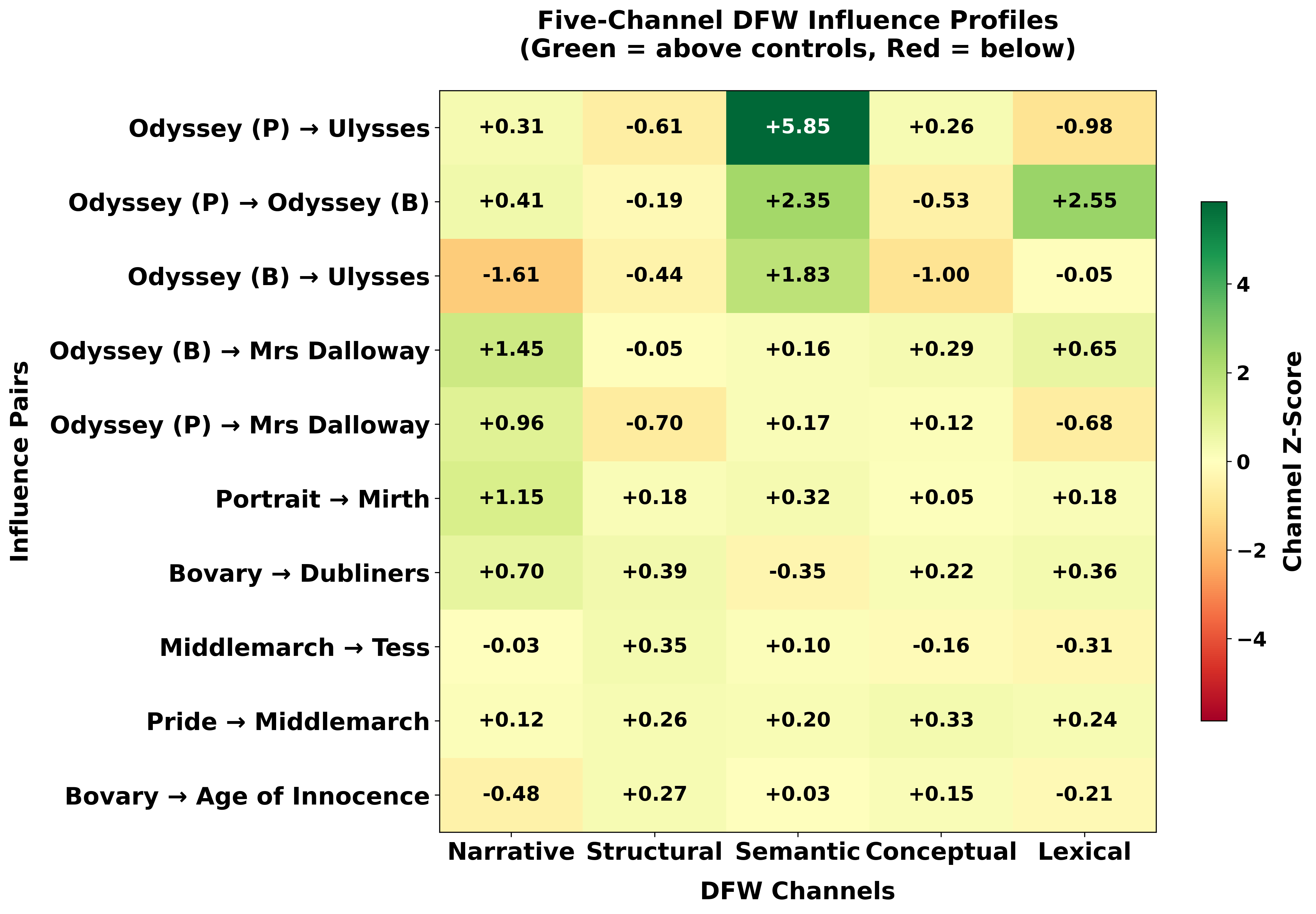}
    \caption{Five-Channel Transformation Profiles}
    \label{fig:channel_profiles}
\end{figure}

\begin{table}
\centering
\small
\begin{tabular}{lcccc}
\hline
Group & $n$ & Mean & Median & Std. Dev. \\
\hline
Reference pairs & 10 & 1.94 & 1.25 & 2.39 \\
Control pairs & 73 & 0.29 & 0.19 & 0.30 \\
\hline
\end{tabular}
\caption{Aggregate channelled transformation scores for reference and control pairs}
\label{tab:aggregate_scores}
\end{table}

%----------------------------------------------------------------------------------%
\section{Experiments and Results}
This section evaluates whether the framework produces distinctive transformation patterns for historically related texts compared to unrelated corpus pairs. Instead of treating influence as binary detection, we examine how retention and divergence are distributed across the five representational dimensions introduced above.

The model is applied to the 10 reference relationships in Table~\ref{tab:influence} and 73 unrelated control pairs constructed from source $\rightarrow$ target orderings. For instance, \textit{Pride and Prejudice} $\rightarrow$ \textit{Mrs Dalloway} is retained as a control because Austen predates Woolf, while the reverse direction is excluded as historically impossible. For each pair, the framework computes channel-specific similarity scores and the aggregate transformation score defined in Equation~\ref{eq:score}.

To assess robustness to segmentation, we run the full pipeline with both chapter-based segmentation and overlapping sliding windows. Reference-control separation is directionally consistent in both settings, but stronger with sliding windows ($r = 0.74$, $p < .001$) than with chapter segmentation ($r = 0.35$, $p = .07$). The remaining analysis therefore reports the sliding-window setting.

The evaluation asks whether historically related works produce stronger transformation signals than controls, and how retention differs across representational layers. Since the model allows selective inheritance, different pairs are not expected to preserve the same kinds of structure. We therefore consider both aggregate score magnitude and the distribution of channel-specific signals.

\subsection{Results}
Table \ref{tab:aggregate_scores} summarizes the transformation scores. Reference pairs obtain a higher mean than controls (1.94 versus 0.29) and show substantially greater variance. A Mann--Whitney $U$ test confirms a significant separation ($U = 635$, $p < 0.001$, rank-biserial $r = 0.74$), indicating that reference pairs generally receive higher scores than controls. The larger variance among reference pairs suggests that influence does not manifest uniformly, but through different retained layers across different relationships.

Figure \ref{fig:channel_profiles} presents the five-channel profiles for the reference pairs. Each row shows standardized channel scores relative to the control distribution, where positive values indicate retention and negative values indicate divergence. The profiles show that historically related pairs do not simply display uniformly higher similarity. Instead, they tend to activate one or two representational dimensions while diverging in others. This supports the central claim that creative transformation is best understood as selective inheritance across representational levels.

\begin{table*}[t]
\centering
\small
\renewcommand{\arraystretch}{1.15}
\begin{tabular}{@{}lccc@{}}
\toprule
Signal & $r$ & AUC & $p$ \\
\midrule
Lexical overlap (text reuse) \cite{allen2011intertextuality,forstall2015modeling} & $-0.04$ & 0.48 & .86 \\
POS-trigram structure & $0.09$ & 0.55 & .64 \\
TF--IDF content similarity & $0.10$ & 0.55 & .60 \\
LDA topic mixture & $0.30$ & 0.65 & .13 \\
Sentence-BERT embedding \cite{burns2021profiling,xing2025modelling} & $0.41$ & 0.71 & .04 \\
\midrule
\textbf{Channelled model (ours)} & $\mathbf{0.74}$ & \textbf{0.87} & $\mathbf{<.001}$ \\
\bottomrule
\end{tabular}
\caption{Reference versus control separation for single-channel baselines and the full channelled model under the sliding-window setting. The evaluation includes 10 reference pairs and 73 control pairs. $r$ is the rank-biserial effect size, AUC is the probability that a reference pair receives a higher score than a control pair, and $p$ is computed using a two-sided Mann--Whitney $U$ test.}
\label{tab:baselines}
\end{table*}
\subsection{Comparison with Single-Channel Baselines}
To assess whether channelled aggregation adds information beyond individual dimensions, we compare the full model with several single-channel baselines. These approximate common approaches to textual relation, including lexical overlap and content similarity used in text-reuse and intertextuality detection \cite{allen2011intertextuality,forstall2015modeling}, as well as dense embedding similarity used in recent embedding-based intertextual analysis \cite{burns2021profiling,xing2025modelling}. For each baseline, per-window scores are aggregated to the pair level and evaluated by their ability to separate reference pairs from controls.

Table~\ref{tab:baselines} shows that no individual channel separates the groups as strongly as the full model. Lexical overlap, POS-trigram structure, and TF--IDF content similarity remain close to chance ($\text{AUC}\le 0.55$). LDA topic mixtures provide a stronger but non-significant signal, while Sentence-BERT embeddings are the strongest single-channel baseline ($r=0.41$, $\text{AUC}=0.71$). The channelled model achieves a larger separation ($r=0.74$, $\text{AUC}=0.87$), suggesting that the relevant signal lies in the configuration of retention and divergence across channels, not in high similarity along one dimension alone.

\subsection{Reference Pair Case Studies}
Several transformation patterns emerge among the reference pairs. The two English translations of the \textit{Odyssey} show strong semantic and lexical retention, reflecting preservation of narrative content under different stylistic realizations. Pope's version renders the epic in heroic couplet verse, while Butler's presents it in unrhymed prose, introducing differences in rhythm, sentence segmentation, and pacing while leaving the underlying semantic structure largely intact.

The Homeric-modernist comparisons show a different configuration. In Joyce's \textit{Ulysses}, which does not reproduce the \textit{Odyssey} at the level of wording or surface style, the strongest signal appears in the semantic channel. This suggests mediated thematic continuity combined with substantial lexical and structural divergence. Since the aggregate score rewards strong retention in one channel together with divergence in others, this pair scores higher than the more uniformly similar translation pair.

The Woolf comparisons are more narrative-oriented. The Butler \textit{Odyssey} $\rightarrow$ \textit{Mrs Dalloway} pair produces a stronger signal than the corresponding Pope comparison, with a peak in the narrative channel. Because Butler's translation uses prose instead of heroic verse, its form is closer to the narrative conventions Woolf transforms. The stronger Butler signal may therefore reflect mediation through translation, corresponding to hypertextual transformation in Genette's terms.

Relationships within later prose traditions produce further configurations. The Henry James $\rightarrow$ Edith Wharton comparison shows elevated narrative and semantic signals, consistent with Wharton's engagement with Jamesian focalization and psychological interiority. The Eliot $\rightarrow$ Hardy comparison peaks in the structural channel, likely reflecting continuity in Victorian realist narration and syntactically complex prose without direct content borrowing.

The Flaubert $\rightarrow$ Joyce comparison is more heterogeneous, with narrative activation and moderate structural and conceptual signals. This fits Flaubert's role in the development of modern narrative technique, especially free indirect discourse and subjective consciousness, which Joyce extends within a modernist context. Other reference pairs, such as \textit{Madame Bovary} $\rightarrow$ \textit{The Age of Innocence}, show weaker but interpretable structural and conceptual continuity, suggesting that influence may also appear as distributed partial retention rather than a single dominant channel.
\begin{table*}[!t]
\centering
\small
\begin{tabular}{lcccccccc}
\hline
Pair & Type & Score & Narr & Struct & Sem & Concept & Lex & Peak \\
\hline
Odyssey (Pope) $\rightarrow$ Ulysses & Influence & 8.19 & 0.31 & -0.61 & 5.85 & 0.26 & -0.98 & Semantic \\
Odyssey (Pope) $\rightarrow$ Odyssey (Butler) & Influence & 3.06 & 0.41 & -0.19 & 2.35 & -0.53 & 2.55 & Lexical \\
Odyssey (Butler) $\rightarrow$ Ulysses & Influence & 2.56 & -1.61 & -0.44 & 1.83 & -1.00 & -0.05 & Semantic \\
The Portrait of a Lady $\rightarrow$ The House of Mirth & Influence & 1.15 & 1.15 & 0.18 & 0.32 & 0.05 & 0.18 & Narrative \\
Middlemarch $\rightarrow$ Tess of the d'Urbervilles & Influence & 0.35 & -0.03 & 0.35 & 0.10 & -0.16 & -0.31 & Structural \\
Pride and Prejudice $\rightarrow$ Middlemarch & Influence & 0.33 & 0.12 & 0.26 & 0.20 & 0.33 & 0.24 & Conceptual \\
\hline
Middlemarch $\rightarrow$ The Portrait of a Lady & Control & 0.60 & 0.60 & 0.02 & 0.09 & 0.08 & 0.25 & Narrative \\
Pride and Prejudice $\rightarrow$ The Voyage Out & Control & 0.81 & 0.81 & 0.39 & 0.17 & 0.50 & 0.69 & Narrative \\
Madame Bovary $\rightarrow$ Tess of the d'Urbervilles & Control & 0.93 & 0.71 & 0.93 & 0.03 & 0.18 & -0.14 & Structural \\
Madame Bovary $\rightarrow$ The House of Mirth & Control & 0.74 & 0.09 & 0.61 & -0.10 & -0.06 & -0.53 & Structural \\
\hline
\end{tabular}
\caption{Representative transformation profiles for influence and control pairs under the chronologically filtered control setting}
\label{tab:results_examples}
\end{table*}
\subsection{Reference Pairs and Controls}
Table \ref{tab:results_examples} presents representative profiles for reference and control comparisons. Reference pairs more often exhibit a pronounced dominant channel, while controls tend to show weaker or more diffuse similarity across channels.

This contrast is important because many controls are not maximally distant comparisons. They often come from similar periods, related prose traditions, or overlapping stylistic environments, and therefore share broad conventions of narration, syntax, and theme. Even against this stronger baseline, reference pairs more often display a clear dominant channel or coherent retained structure.

The full ranking across all 83 pairs shows the same tendency: the six highest transformation scores are all reference pairs, and seven of the ten reference pairs fall within the top 20. The three lower-ranking reference pairs have flatter profiles, suggesting that some realist influence may be distributed weakly across several channels. Overall, the comparison supports the view that creative transformation operates through selective retention and divergence, not uniform similarity.

%----------------------------------------------------------------------------------%
\subsection{Limitations}
The present study is limited by the size and composition of its corpus, which focuses on a small set of canonical works from the eighteenth to early twentieth centuries. It should therefore be understood as an exploratory evaluation setting for developing the model, not as a representative sample of literary creativity in general. Other forms of influence, including diffuse genre conventions, multilingual transmission, unconscious stylistic inheritance, or mediation through broader cultural movements, may require larger datasets and additional representational channels.

The reference pairs are based on relationships discussed in literary scholarship rather than formally verifiable ground truth, and should be interpreted as historically plausible cases of transformation, not definitive instances of influence. Likewise, the model's representational channels are simplified proxies for complex literary phenomena: narrative structure, conceptual domains, and stylistic organization cannot be fully captured by the features used here. The resulting profiles should therefore be read as interpretable evidence of structural relation and as prompts for further literary analysis, not as replacements for close reading. Finally, because the framework assumes a plausible precursor, it is less applicable to works whose originality lies mainly in departing from a broader field; such cases may be better characterized by distributional novelty measures, making genealogical transformation and corpus-level novelty complementary perspectives.

%----------------------------------------------------------------------------------%
\section{Conclusion}
This paper introduced a multi-channel framework for modelling creative transformation between literary texts. Instead of treating creativity as deviation within a corpus, the proposed approach analyses how representational structures are selectively preserved or transformed across lexical, semantic, conceptual, structural, and narrative levels.

Applying the framework to a corpus of canonical literary works shows that historically related texts tend to exhibit distinctive transformation profiles. Different relationships preserve different layers of structure, ranging from narrative organization in closely related prose traditions to semantic or conceptual continuity in more distant reinterpretations such as modernist engagements with classical epic. These findings support the view that literary creativity often operates through selective inheritance, with similarity concentrated in specific representational layers instead of appearing uniformly across texts.

More broadly, the results suggest that modelling influence as a multi-layered relation between specific works offers a complementary perspective to existing distribution-based approaches to computational creativity. By decomposing textual comparison across representational channels, the framework provides a way to analyse how creative works simultaneously inherit and transform earlier structures.

The framework may also be useful for evaluating creative text generation. For example, systems prompted to imitate an author or transform an existing narrative may reproduce surface style without preserving deeper semantic, conceptual, or narrative structure. A multi-channel transformation profile could help distinguish shallow stylistic imitation from more substantive transformation of source material.

Future work may extend this approach to larger corpora, additional literary traditions, multilingual settings, and other cultural domains where creativity emerges through structured transformation of prior artifacts.
%----------------------------------------------------------------------------------%
\section{Acknowledgements}
The authors thank the Romanian writer Ștefan Agopian, whose reflections on literature and the creative process inspired the central idea of this work. They are also grateful to Professor Andra Băltoiu, whose suggestions on stylistic influence and creative transformation helped shape the initial direction of this work.

This work was supported by Ministry of Research, Innovation and Digitization, CNCS-UEFISCDI, project SIROLA, number PN-IV-P1-PCE-20231701, within PNCDI IV.

 %----------------------------------------------------------------------------------%
\bibliographystyle{iccc}
\bibliography{iccc}

\end{document}